\documentclass[letterpaper, 10 pt, conference]{ieeeconf}

\IEEEoverridecommandlockouts                              

\usepackage[l2tabu,orthodox]{nag}

\usepackage[
    backend=biber,
    bibencoding=utf8,
    style=ieee,
    sorting=none,
    natbib=true,
    doi=false,
    isbn=false,
    url=false,
    eprint=false,
    maxcitenames=1,
    mincitenames=1,
]{biblatex}

\usepackage[draft,pdftex,colorlinks]{hyperref}

\usepackage[printonlyused]{acronym}

\usepackage{siunitx}
\usepackage[all]{nowidow}

\usepackage[dvipsnames]{xcolor}

\usepackage{lipsum}

\usepackage[pdftex]{graphicx}

\usepackage{epstopdf}

\usepackage{import}

\graphicspath{{./latexGoodPractices/}}

\usepackage{booktabs}

\usepackage{tabu}
\usepackage{makecell}
\usepackage{float}

\usepackage{amssymb,amsfonts,amsmath,amscd}

\usepackage{cleveref}
\creflabelformat{equation}{#2#1#3}

\usepackage{bm}

\newcommand{\norm}[1]{\left\Vert#1\right\Vert}

\newcommand{\bbm}{\begin{bmatrix}}
        \newcommand{\ebm}{\end{bmatrix}}

\usepackage[normalem]{ulem} 

\makeatletter
\usepackage[utf8]{inputenc}
\usepackage{commath}
\usepackage{mathtools}
\usepackage{mathabx}
\usepackage{siunitx}
\usepackage[ruled,vlined]{algorithm2e}
\usepackage[lofdepth,lotdepth]{subfig}
\usepackage{tikz}

\let\oldtheequation\theequation
\renewcommand\tagform@[1]{\maketag@@@{\ignorespaces#1\unskip\@@italiccorr}}
\renewcommand\theequation{(\oldtheequation)}
\makeatother

\usepackage[belowskip=0pt,aboveskip=1pt]{caption}
\graphicspath{{media/}}

\newcommand{\ie}{i.e., }
\newcommand{\eg}{e.g., }

\newcommand{\se}{\mathfrak{se}}

\DeclareMathOperator*{\argmin}{arg\,min}

\usepackage{amssymb}
\usepackage{pifont}

\acrodef{ICP}{Iterative Closest Point}
\acrodef{GP}{Gaussian Process}
\acrodefplural{GP}{Gaussian Processes}
\acrodef{RTS}{Robotic Total Station}
\acrodefplural{RTS}{Robotic Total Stations}
\acrodef{DOF}{Degrees Of Freedom}
\acrodef{GNSS}{Global Navigation Satellite System}
\acrodef{RTK}{Real Time Kinematics}
\acrodef{UGV}{Unmanned Ground Vehicle}
\acrodef{UAV}{Unmanned Aerial Vehicle}
\acrodef{IMU}{Inertial Measurement Unit}
\acrodef{MC}{Monte Carlo}
\acrodef{EDM}{Electronic optical Distance Measurement}
\acrodef{GUM}{Guide to the expression of Uncertainty in Measurement}
\acrodef{SLAM}{Simultaneous Localization and Mapping}
\acrodef{STEAM}{Simultaneous Trajectory Estimation And Mapping}
\acrodef{GCP}{Ground Control Point}
\acrodefplural{GCP}{Ground Control Points}

\DeclareUnicodeCharacter{0308}{¨} 

\begin{document}
\title{\LARGE \textbf{Uncertainty analysis for accurate ground truth trajectories with robotic total stations}}

\author{Maxime Vaidis$^{1}$, William Dubois, Effie Daum, Damien LaRocque, François Pomerleau$^{1}$
  \thanks{$^{1}$Northern Robotics Laboratory, Université Laval, Québec City, Canada,
    {\texttt{\small{$\{$maxime.vaidis, francois.pomerleau$\}$ @norlab.ulaval.ca}}}}%
}

\linepenalty=3000
\addtolength{\textfloatsep}{-0.1in}

\maketitle
\thispagestyle{empty}
\pagestyle{empty}


\begin{abstract}
In the context of robotics, accurate ground truth positioning is essential for the development of \ac{SLAM} and control algorithms.
\acp{RTS} provide accurate and precise reference positions in different types of outdoor environments, especially when compared to the limited accuracy of \ac{GNSS} in cluttered areas.
Three \acp{RTS} give the possibility to obtain the six-\ac{DOF} reference pose of a robotic platform.
However, the uncertainty of every pose is rarely computed for trajectory evaluation.
As evaluation algorithms are getting increasingly precise, it becomes crucial to take into account this uncertainty.
We propose a method to compute this six-\ac{DOF} uncertainty from the fusion of three \acp{RTS} based on \ac{MC} methods.
This solution relies on point-to-point minimization to propagate the noise of \acp{RTS} on the pose of the robotic platform.
Five main noise sources are identified to model this uncertainty: noise inherent to the instrument, tilt noise, atmospheric factors, time synchronization noise, and extrinsic calibration noise.
Based on extensive experimental work, we compare the impact of each noise source on the prism uncertainty and the final estimated pose.
Tested on more than \SI{50}{km} of trajectories, our comparison highlighted the importance of the calibration noise and the measurement distance, which should be ideally under \SI{75}{\metre}.
Moreover, it has been noted that the uncertainty on the pose of the robot is not prominently affected by one particular noise source, compared to the others.
\end{abstract}



\acresetall
\section{Introduction}

\begin{figure}[t]
\centering
\includegraphics[width=\linewidth]{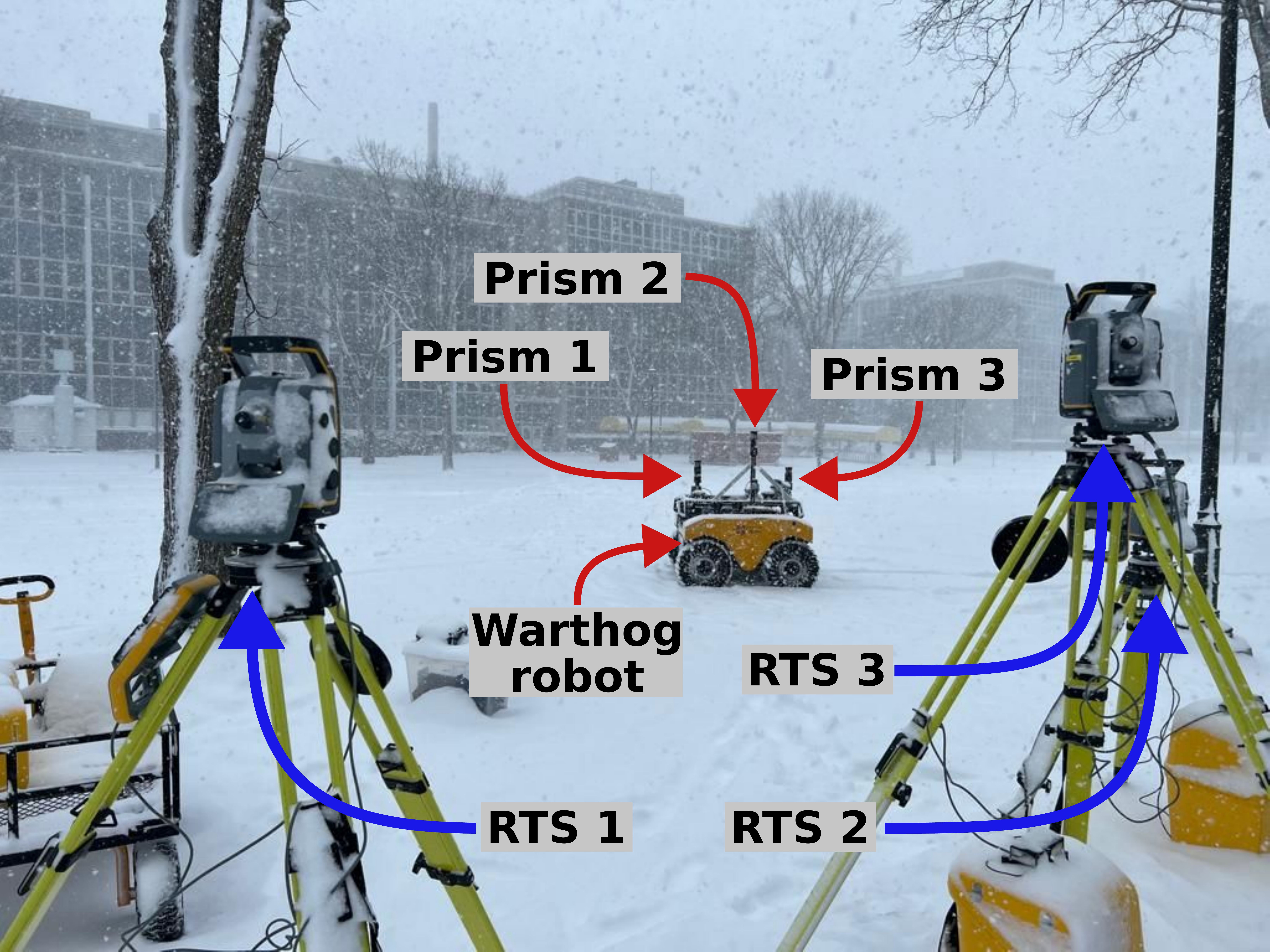}
\caption{Setup used to record a reference trajectory during a snowstorm.
Three \acp{RTS} are each tracking a specific active prism, all mounted on a Clearpath Warthog robotic platform.}~\label{fig:fig-intro}
\vspace{-9mm}
\end{figure}

In mobile robotics, the current development of mapping and control algorithms heavily relies on datasets~\cite{Hilti2022}. 
The performance of these algorithms is evaluated by comparing the different poses with a reference trajectory.
In outdoor environments, \acp{RTS} provide the highest accuracy by measuring reference trajectories with uncertainty on the position in the range of millimeters~\cite{Kalin2022}. 
Coming from the field of surveying, a \emph{total station} is an optic-based measurement instrument that can be precisely aimed at a given prismatic retro-reflector (\ie simply called \emph{prism} in the remainder of this article).
A total station is \emph{robotic} when it can automatically track a prism, while this prism is in motion. 
The position of the prism is computed in the local coordinate system of the \ac{RTS}, according to the horizontal and vertical angles, along with the range between the \ac{RTS} and the measured prism.
With three prisms or more attached to a robotic platform, it is possible to compute its six-\ac{DOF} pose through manual static measurement~\cite{Pomerleau2012} or through the use of multiple \acp{RTS} continuously tracking three \emph{active prisms} rigidly mounted on the same platform~\cite{Vaidis2021}.
Active prisms are recently available off the shelf and provide a unique light signature for automatic target identification by \acp{RTS}.
Each prism is tracked by its own assigned \ac{RTS}, as shown in~\autoref{fig:fig-intro}.
The distance between a \ac{RTS} and its prism is determined by \ac{EDM}, which is greatly impacted by weather conditions~\cite{Rueger2002}.

Yet, to be useful in autonomous navigation research, evaluation protocols need to be used in a variety of conditions and environments, such as snowfalls~\cite{Pomerleau2023}, which can increase the uncertainty of ground truth trajectories.
In addition to measurement noise inherent to a single \ac{RTS}, using multiple \acp{RTS} involves time synchronization and extrinsic calibration to fuse the data of all \acp{RTS} in a common frame~\cite{Vaidis2022}.
Both this synchronization and calibration carry noises and uncertainties that must be studied.
Uncertainty analysis is not usually part of the \ac{SLAM} algorithms evaluation pipelines, as the most common metric for comparing trajectories is the Euclidean distance.
However, as current algorithms are developed with the intention to be accurate, an evaluation that does not consider uncertainties will lead to biased results. 
Moreover, uncertainty estimations for ground truth trajectories are missing in state-of-the-art outdoor \ac{SLAM} datasets.
Two factors could explain this:
\textbf{1)}~Since ground truth trajectory noises are considered negligible, uncertainty is not computed.
\textbf{2)}~It can be complex to model the uncertainty of a reference system, such as \acp{GNSS}. 
Uncertainty models were developed for \acp{RTS}~\cite{Ulrich2013}, but they were never used for trajectory evaluations on mobile robots driven in outdoor environments.

This work is based on our previous research to develop an \ac{RTS} setup for trajectory evaluations~\cite{Vaidis2021, Vaidis2022}.
In this paper, we propose a method to model \ac{RTS} uncertainties with the objective to better compare six-\ac{DOF} trajectories.
For that purpose, we carried out a \ac{MC} method that includes five different sources of uncertainty relative to multiple \ac{RTS} measurements in outdoor environments.
These uncertainties are then interpolated over time by a \ac{GP} and propagated to the reference pose of a robot by using another \ac{MC} method.
A detailed qualitative analysis of the different noise sources is presented, as well as their impact on the final resulting six-\ac{DOF} trajectories.
The experimental data used to compute the results was gathered during a whole year of deployments, with over \SI{50}{\km} of recorded trajectories in different weather conditions and environments.
Both our source code and our dataset are freely available in our \texttt{RTS\_Project} repository.\footnote{\url{https://github.com/norlab-ulaval/RTS_project}}

\section{Related work}
\label{sec:related_work}

We first describe the current uses of \acp{RTS} to obtain reference trajectories for mobile robotics.
Then, we present different studies of \ac{RTS}-related uncertainties and we expose different methods that are used in the state of the art to model and propagate uncertainties.
Finally, we address the use of these methods in mobile robotics and we discuss their properties.

\ac{RTS}-based positioning systems are quite common in mobile robotics.
The number of \acp{RTS} in an experimental setup is determined by the number of prisms that can be handled by the robotic platform, as well as the number of \acp{DOF} in the desired resulting trajectory.
A single \ac{RTS} was used to acquire the three-\ac{DOF} position of a prism mounted on different robotic platforms, such as a planetary rover~\cite{Lemus2014}, a tracked robot~\cite{Kubelka2015}, an unmanned surface vessel~\cite{Hitz2015}, a skid steered robot~\cite{MacTavish2018}, and a \ac{UAV}~\cite{Schmuck2019}.
It is possible to reduce the uncertainty of the reference position by adding a second \ac{RTS} to track the same prism. 
\citet{Reitbauer2020} have used a second \ac{RTS} to follow two different prisms on a compost turner, enabling the measurement of four \acp{DOF} on the platform (\ie the position and the yaw angle).
To obtain the full pose reference of a static robotic platform, it is possible to manually measure three prisms with a single \ac{RTS}~\cite{Pomerleau2012}.
Furthermore, this setup provides a quantitative way to analyze uncertainty through inter-prism distances.
These distances can be compared with values that were accurately determined in a controlled environment.
For a moving platform, \citet{Vaidis2021} developed the first method to compute and interpolate six-\ac{DOF} poses of a robot, with the measurements of three \acp{RTS}.
In this paper, we build on this method by providing a continuous six-\ac{DOF} pose uncertainty model that relies on \acp{RTS} measurements.

Many noise sources can be modeled and used to estimate the uncertainty of a \ac{RTS}'s measurement.
Each noise model has an impact on different parts of a \ac{RTS} processing pipeline, from the raw measurements of the \ac{RTS} to the estimated Cartesian position of a prism.
Most uncertainty sources are directly related to the devices (measuring instruments and prisms).
Distances and angles uncertainties can be estimated with manufacturer's specifications, or with experimental results, done both in laboratories. 
Outside these controlled environments, the atmospheric factors (\eg temperature, pressure, and humidity) need to be considered, due to \ac{EDM} sensitivity~\cite{Rueger2002}.
As such, a variation on \SI{1}{\celsius} can lead to an error of \SI{0.2}{\milli\metre} on a measured distance of \SI{200}{\metre}~\cite{Rodriguez2021}.
The noise of a \acl{RTS}'s electronic compensator can be estimated through manufacturer specifications, yet the associated uncertainty is often disregarded when conducting precise surveying~\cite{Werner2017}.
Moreover, time synchronization errors and uncertainties can occur in the communication between a \ac{RTS} and an external controller or data acquisition system~\cite{Thalmann2021}.
When using multiple \acp{RTS}, the accuracy of the extrinsic calibration between all \acp{RTS} influences the accuracy of the estimated prism positions.
\citet{Vaidis2022} implemented a pipeline to filter outliers on \ac{RTS} data and proposed an extrinsic calibration method that corrects the error on the poses, yet uncertainty remained.
Finally, a moving target creates some additional uncertainties that are difficult to quantify.
This noise comes from the limitations of the \acp{RTS} angular tracking system, especially at high prism speeds and accelerations~\cite{Morse2015}.
When using multiple prisms, the inter-prism distances can be used to filter imprecise results with a threshold on prism speeds~\cite{Vaidis2021}.
This paper examines all these sources of uncertainty, to model the global uncertainty of each \ac{RTS} measurement under different atmospheric conditions.

There are two main ways to model the total uncertainty of a \ac{RTS}, based on the aforementioned sources of uncertainty: either with an approach that is based on the \acfi{GUM}, or with \ac{MC} simulations.
The \ac{GUM}~\cite{GUM} divides uncertainties into two types, between those obtained from statistical analysis on a series of observations (defined as \emph{Type A}), and those expressed by average manufacturer-specified or user-defined values (defined as \emph{Type B}).
With the \ac{GUM} method, an uncertainty budget of a \ac{RTS} allows one to express the total uncertainty of this \ac{RTS} as an isotropic noise~\cite{SIAUDINYTE2016}.
This method works well for noise sources that can be linearized, but it can be complex to implement for non-linear noise, such as weather conditions.
For this reason, \ac{MC} simulations are widely used to determine the uncertainty of \ac{RTS}, whether for simple models~\cite{Fumin2008}, or very complex models taking into account non-linear noise, such as atmospheric factors~\cite{Ulrich2013}.
Moreover, the resulting uncertainty is modeled as anisotropic.
Generally, a \ac{MC} method relies on between \(10^3\) to \(10^5\) samples to have coherent generated results, making this method computationally greedy for large datasets~\cite{Ferson1996}.
Both of these methods give an estimate of a prism's position uncertainty, yet it is unsuitable for mobile robotics when the uncertainty is propagated into the reference frame of a robotic platform.
Several algorithms exist to propagate uncertainty in a system.
An Unscented Kalman filter can be used to estimate the resulting noise~\cite{Hu2020}.
The Unscented transform method has been carried out to tackle computational resources issues, with as accurate results for uncertainty estimation as with \ac{MC}~\cite{Dailys2020}.
This method is based on the key idea that it should be easier to approximate a probability distribution than to approximate an arbitrary nonlinear function.
Yet, the Unscented transform is only applied to points of a specific covariance distribution at a time.
Since three-\acp{RTS} positioning systems give three different covariance distributions, other methods that can process multiple distributions are more appropriate.
Other studies have used Lie Algebra to link and interpolate the pose of a system to its uncertainty.
\citet{Barfoot2014} formalized ways to work with noise in $\se(3)$ and applied them to propagate the noise from a camera over a trajectory.
\citet{Anderson2015} developed a library called \ac{STEAM} that uses \acp{GP} to interpolate the covariance matrix of a system for nonlinear optimization problems with continuous-time components. 
In this article, we combine the research of \citet{Ulrich2013} and \citet{Anderson2015} to propagate the uncertainty to the pose of a robotic platform.

\section{Theory}
\label{sec:theory}


We first present our approach for modeling uncertainty of \ac{RTS} measurements with the \ac{MC} method.
Then, we show how we interpolate data with \acp{GP} for prism measurement uncertainties.
Next, we describe how we use another \acl{MC} method to propagate uncertainty from interpolated prism measurements to six-\ac{DOF} vehicle poses.

\subsection{\acl{RTS} noise models}
\label{sec:theoryA}
As highlighted in \autoref{sec:related_work}, the uncertainty on the measurement from a \ac{RTS} is impacted by different noise sources.
Each noise source can be defined with a stochastic model, hence the possibility to use a \ac{MC} method to estimate the resulting combination of all sources of uncertainty on a single \ac{RTS} measurement.
We defined a trajectory $\mathcal{P}^i$ in the frame $\mathcal{F}^i$, where $i \in \{1,2,3\}$ is the index of a single \ac{RTS}, as a set of normalized homogeneous prism coordinate measurements $\{\bm{p}_1^i, \dots, \bm{p}_{n_i}^i\}$ such that $\bm{p}_k^i$ is the $k^{\text{th}}$ measurement of $\mathcal{P}^i$ with $k \in \{1,n_i\}$ and $n_i\in \mathbb{N^*}$ is the number of measurements for the $i$-th \ac{RTS}.
By merging all different kinds of noises with a \ac{MC} method, we are able to determine the spatial covariance $\bm{\Sigma}^i_k$ around each measurement $\bm{p}_k^i$.
In the following paragraphs, we define five uncertainty models with their parameters that were used to describe the noise encountered during our deployments with multiple \acp{RTS}.

\textbf{\ac{RTS} instrument noises --}
These noise sources are directly coming from multiple errors in the instrument calibration, namely the vertical collimation error, the centering error, the horizontal collimation error, and the eccentricity error.
They alter the raw measurements given by the \ac{RTS}, namely the distance $\rho$, and both the horizontal and vertical angular values, $\phi$ and $\theta$, which are used to compute prism coordinates.
Their standard deviations $\sigma_{\rho}$, $\sigma_{\phi}$ and $\sigma_{\theta}$, respectively for the distance, horizontal and vertical deviation, are given by manufacturers in the instrument specifications.
Then, errors on measurements can be represented by a zero-mean normal distribution, respectively $\epsilon_{\rho} \sim \mathcal{N}(0,\sigma_{\rho})$, $\epsilon_{\phi} \sim \mathcal{N}(0,\sigma_{\phi})$ and $\epsilon_{\theta} \sim \mathcal{N}(0,\sigma_{\theta})$.

\textbf{Tilt compensator --}
Modern \ac{RTS} are equipped with an electronic angular compensator that allows the instrument to correct pitch and roll values, with its estimated gravity vector.
This compensator has an inherent noise $\epsilon_{tilt}$ represented by a zero-mean normal distribution $\epsilon_{tilt} \sim \mathcal{N}(0,\sigma_{tilt})$ as described by \citet{Werner2017}.

\textbf{Atmospheric factors and weather --}
Since distance measurements are taken with \ac{EDM}, they are subject to the influence of atmospheric factors, specifically temperature \(T\), pressure \(P\), and humidity \(H\)~\cite{Ulrich2013}.
These atmospheric factors are represented by uniform distributions $\epsilon_T$, $\epsilon_P$, and $\epsilon_H$.
According to equations proposed by \citet{Rueger2002}, these uniform distributions will lead to the estimation of a correction factor $\alpha$ (expressed in \si{ppm}) to rectify a measured distance $\rho$.
The aforementioned measurement noise sources ($\epsilon_{\rho}$, $\epsilon_{\phi}$, $\epsilon_{\theta}$, $\epsilon_{tilt}$ and the correction $\alpha$) are combined to include uncertainties to raw \ac{RTS} measurements:
\begin{align}
    \widehat{\rho}   & = (\rho + \epsilon_{\rho})(1+\alpha) \label{eq:raw-rho},\\
    \widehat{\theta} & = \theta + \epsilon_{\theta} + \epsilon_{tilt} \label{eq:raw-theta},\\
    \widehat{\phi}   & = \phi + \epsilon_{\phi} + \epsilon_{tilt}\cot(\widehat{\theta}). \label{eq:raw-phi}
\end{align}

\textbf{Time synchronization --}
Data acquisition made by several \acp{RTS} leads to a time synchronization error $\epsilon_{t_s}$, expressed in seconds.
The resulting uncertainty $\bm{\epsilon_t}$ alters the Cartesian coordinates of a prism and is related to the velocity $\bm{v}_k^i$ at which it moves.
According to \citet{Ulrich2013}, this time synchronization uncertainty follows a normal distribution $\bm{\epsilon_t} \sim \mathcal{N}(\bm{\mu_t},\bm{\sigma_{t}})$ that depends on the time synchronization error $\epsilon_{t_s} \sim \mathcal{N}(\mu_{t_s},\sigma_{t_s})$ and the prism velocity $\bm{v}_k^i \sim \mathcal{N}(\bm{\mu_{v}},\bm{\sigma_{v}})$:
\begin{align}
    \bm{\mu_{t}} & = \mu_{t_s}\bm{{\mu_v}}\\
    \bm{\sigma_{t}^2} & = \mu_{t_s}^2\bm{\sigma_v^2}
    +\sigma_{t_s}^2\bm{{\mu_v}^2},
\end{align}
where $\mu_{t_s}$ represents the mean time synchronization error, $\sigma_{t_s}$ its standard deviation and $\bm{\mu_v^2} = \begin{bmatrix}\mu_{v_x}^2 & \mu_{v_y}^2 & \mu_{v_z}^2 \end{bmatrix}^{\intercal}$ is the square of the average prism velocity vector with a covariance $\bm{\sigma_{v}^2}=\begin{bmatrix}\sigma_{v_x}^2 & \sigma_{v_y}^2 & \sigma_{v_z}^2    \end{bmatrix}^{\intercal}$.
The prisms' velocities are estimated by differentiating the prism Cartesian coordinates with respect to time, by considering computed uncertainties from \cref{eq:raw-rho,eq:raw-theta,eq:raw-phi}, such that:
\begin{gather}
    \bm{p}_k^i =
    \begin{bmatrix}\widehat{\rho_k}\sin{\widehat{\phi_k}}\cos{\widehat{\theta_k}} & \widehat{\rho_k}\sin{\widehat{\phi_k}}\sin{\widehat{\theta_k}} & \widehat{\rho_k}\cos{\widehat{\phi_k}}\end{bmatrix}^{\intercal} \label{eq:point}\\
    \bm{v}_k^i = \frac{\bm{p}_{k+1}^i-\bm{p}_k^i}{t_{k+1}-t_k}. \label{eq:speed}
\end{gather}
The values of $\bm{\mu_{v}}$ and $\bm{\sigma_{v}}$ can be estimated for each prism position by applying a \ac{MC} method to prism speeds given by \cref{eq:point,eq:speed}.
A time synchronization error $\epsilon_{t_s}$ can be estimated over a span of time, by taking into account the rate at which the external system's clock diverges from the \ac{RTS}'s clock.
The time synchronization method presented by \citet{Vaidis2021} yields time drift measurements, equal to the worst drift at the end of every time synchronization period (\SI{5}{\minute} in the current case).
When these measurements are recorded for all deployments on the field, they form a distribution of drifts, among which we can statistically determine the values of $\mu_{t_s}$ and $\sigma_{t_s}$.
The estimated error $\bm{\epsilon_t}$ is then added to each prism position.


\textbf{Extrinsic calibration --}
This calibration determines the rigid transformations ${}^W_iT$ between the reference frame $\mathcal{F}^W$ and the frame $\mathcal{F}^i$ of each \ac{RTS}.
Our previous work~\cite{Vaidis2022} exposed many extrinsic calibration methods, including the static \acp{GCP} calibration, that will be used in this paper.
As defined in~\cite{Vaidis2022}, a \ac{GCP} is a position measured on the ground with a static prism used as a target.
A number $n$ of \acp{GCP} is measured in an environment with all \acp{RTS}.
The outcome of this calibration will have some noise, as the earlier-mentioned uncertainties on single measurements propagate in the process.
As the extrinsic calibration is complex to model, its uncertainty is estimated by applying another \ac{MC} method to each \ac{GCP}.
The instrument noises, tilt compensator noise, and atmospheric factors are considered for this \ac{MC} method.
Time synchronization was not included due to the static nature of \acp{GCP}, and because extrinsic calibration yields results that are independent of time.
An extrinsic calibration is computed for each set of \ac{MC} samples for each \acp{GCP}.
The resulting rigid transformation is then applied to each prism trajectory as $\mathcal{Q}^i = {}^W_i\widehat{T}\;\mathcal{P}^i$, where $\mathcal{Q}^i=\{\bm{q}_1^i, \dots, \bm{q}_{n_i}^i\}$ represents the prism trajectory of the $i$-th \ac{RTS} in the global frame $\mathcal{F}^W$.
The extrinsic calibration uncertainty is estimated from the distribution of the points along those trajectories.

Applying all the noises on \ac{RTS} measurements with a \ac{MC} method enables us to estimate the covariance matrix $\bm{\Sigma}^i_k$ of each measurement $q_k^i$ in $\mathcal{Q}^i$.
In the rest of this paper, the frame of the first \ac{RTS} $\mathcal{F}^1$ is chosen as the global frame.


\subsection{Prism position uncertainty interpolation}~\label{sec:theoryB}

The aim of trajectory evaluation for \ac{SLAM} is to compare a reference trajectory with a six-\ac{DOF} trajectory of a robotic platform computed from various sensors (\eg lidar, \ac{IMU}, \ac{GNSS}), usually defined with different acquisition rates.
Therefore, interpolation is required to synchronize both trajectories.
A \ac{GP} regression approach is chosen for this state estimation, as proposed by~\citet{Barfoot2015}.
This allows us to represent the prism trajectories in continuous time in order to query position values for a desired timestamp.
To guarantee a unique solution, we modelize a prior distribution of the potential trajectories, as a unidimensional \ac{GP}, such that:
\begin{align}
    \bm{x}(t) \sim \mathcal{GP}(\bm{\check x}(t), \bm{\check P}(t,t^\prime)) \text{,} \quad t_0 < t,t^\prime \\
    \bm{y_n} = \bm{g}(\bm{x}(t_n)) + \bm{n_n} \text{,} \quad t_1< t_n <t_N,
\end{align}
where $\bm{x}(t)$ represents the normalized homogeneous prism coordinates at time $t$, $\bm{\check x}(t)$ is the prior mean function, $\bm{\check P}(t, t^\prime) $ is the prior covariance function between two different times $t$ and $t^\prime$, $\bm{y_n}$ are measurements, $\bm{n_n} \sim \mathcal{N}(\bm{0}, \bm{\Sigma})$ is a Gaussian measurement noise, $\bm{g}(\cdot)$ is a nonlinear measurement model, and \(\left\{t_1, \dots, t_n, \dots, t_N\right\}\) is a sequence of measurement times.

In this paper, $\bm{y_n}$ are the measurements in $\mathcal{Q}^i$, the covariance $\bm{\Sigma}$ of $\bm{n_n}$ is estimated by the \ac{MC} method presented in \autoref{sec:theoryA} (\ie $\bm{\Sigma}^i_k$), $\bm{g}(\cdot)$ is the non-linear process of having the measurements in $\mathcal{Q}^i$ by the \ac{RTS} and $t_n$ are the timestamps of $\bm{q}_k^i$.
The interpolated results $\widehat{\mathcal{Q}}^i$, which are expressed by $\bm{x}(t)$, are computed by the \ac{STEAM} library~\cite{Anderson2015} for desired query times.
As a result, each estimated point $\widehat{\bm{q}}^i_j$ in $\widehat{\mathcal{Q}}^i$ has its associated estimated covariance matrix $\widehat{\bm{\Sigma}}^i_j$ coming from the \ac{GP} interpolation, where $j \in \{1,J\}$ is the interpolated prism positions index, and $J \in \mathbb{N^*}$ is the total number of interpolated prism positions.


\subsection{Uncertainty propagation to ground truth trajectory}~\label{sec:theoryC}

With only one \ac{RTS}, the uncertainty $\widehat{\bm{\Sigma}}^i$ on the reference trajectory can be exploited right away to evaluate the reference position of a robotic platform.
However, the robotic platform's pose needs to be evaluated in six-\ac{DOF}.
With three \acp{RTS}, it is possible to obtain the reference pose by doing a point-to-point minimization between the triplets of measured prism coordinates, and the reference triplets measured in laboratory~\cite{Vaidis2021,Vaidis2022}.
Prism uncertainties can be propagated by applying a \ac{MC} sampling with this point-to-point minimization.

\begin{figure}[htbp]
    \centering
    \includegraphics[width=\linewidth]{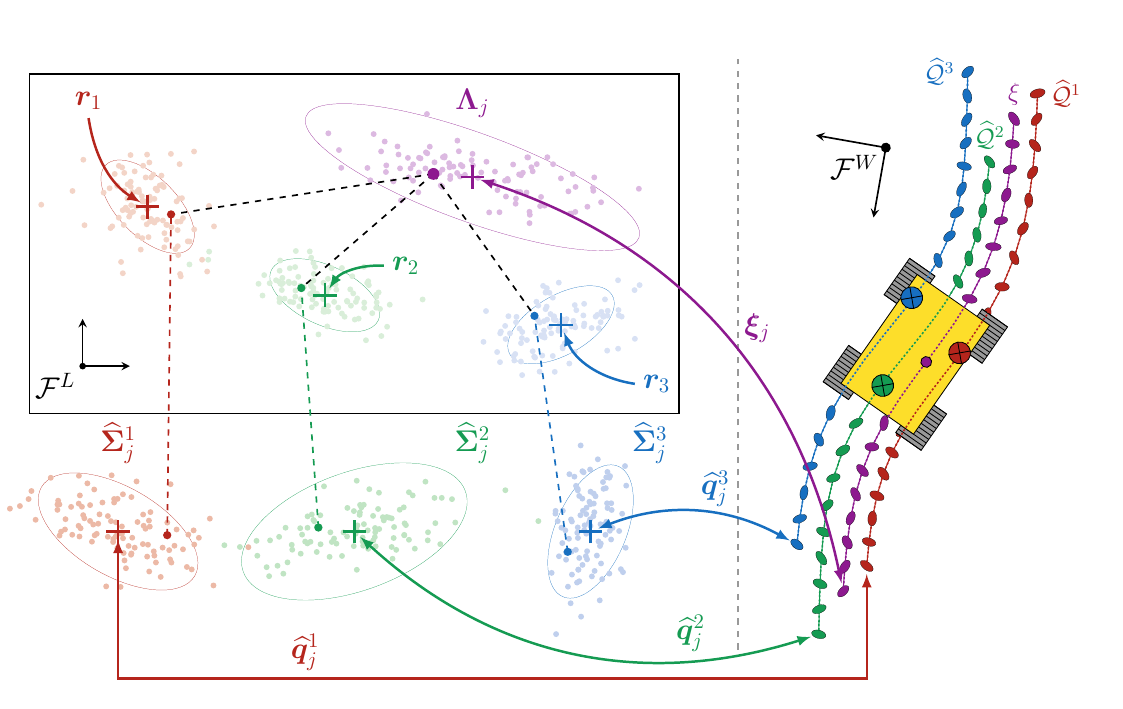}
    \caption{Visualization of error propagation of the \acf{MC} method applied with a point-to-point minimization.
    Each triplet of interpolated prism measurements $\widehat{\bm{q}}^i_j$ of trajectories $\widehat{\mathcal{Q}^i}$ in the world frame $\mathcal{F}^W$ are denoted by crosses enclosed within ellipses representing their corresponding covariances $\widehat{\bm{\Sigma}}^i_j$.
    The points $\bm{r}_1$, $\bm{r}_2$, and $\bm{r}_3$ are the reference prism positions taken in laboratories, along with their covariance.
    A point-to-point minimization minimizes the distance between samples of the same color, in both kinds of distributions.
    The result of this minimization is the estimated vehicle pose $\bm{\xi}_j$ along with its covariance $\bm{\Lambda}_j$.
    }~\label{fig:results_steam}
\end{figure}

After the \ac{GP} interpolation, a setup of three \ac{RTS} yields a set of three paths $\{\widehat{\mathcal{Q}}^1, \widehat{\mathcal{Q}}^2, \widehat{\mathcal{Q}}^3\}$ of interpolated measurements, with their respective covariance $\{\widehat{\bm{\Sigma}}^1, \widehat{\bm{\Sigma}}^2, \widehat{\bm{\Sigma}}^3\}$.
We define \(\widehat{\mathcal{Q}}_j\) as the \(j\)-th triplet of interpolated prism positions with its corresponding triplet \(\widehat{\bm{\Sigma}}_j\) of covariances, such that \(\widehat{\mathcal{Q}}_j = \{\widehat{\bm{q}}^1_j,\widehat{\bm{q}}^2_j,\widehat{\bm{q}}^3_j\}\), and \(\widehat{\bm{\Sigma}}_j = \{\widehat{\bm{\Sigma}}^1_j, \widehat{\bm{\Sigma}}^2_j, \widehat{\bm{\Sigma}}^3_j\}\), as shown in \autoref{fig:results_steam}.
A reference triplet \(\mathcal{R}\) contains normalized homogeneous points $\bm{r}_i$, where $\mathcal{R} = \{\bm{r}_1,\bm{r}_2,\bm{r}_3\}$ with a covariance $\bm{U}_i$ associated to each point $\bm{r}_i$.
These reference points are defined in another world frame $\mathcal{F}^L$ and were statically estimated with measurements from a single \ac{RTS} after each deployment.
To apply a \ac{MC} method, we sample $M$ points \(\left\{s_1^i, \dots, s_m^i, \dots, s_M^i\right\}\) from every Gaussian distribution $\mathcal{N}(\widehat{\bm{q}}^i_j, \widehat{\bm{\Sigma}}^i_j)$, and $M$ points from the Gaussian distribution $\mathcal{N}(\bm{r}_i, \bm{U}_i)$ defined for the triplet of points in $\mathcal{R}$ along with their covariances $\bm{U}_i$.


For every sample \(s_m \in \{s_1,\dots,s_M\}\), we applied the point-to-point minimization with:
\begin{equation}
    {}^1_L\widehat{\bm{T}}_{j,m} =
    \argmin_{\bm{T}} \sum_{i=1}^{3}
    \norm{\widehat{\bm{q}}^{i}_j - \bm{T}{\bm{r}}_{i}}^2_2, 
    \label{eq:point-to-point-minimization}
\end{equation}
where ${}^1_L\widehat{\bm{T}}_{j,m} \in \mathbb{R}^{4\times4}$ is the resulting rigid transformation of the \ac{MC} method between the frame $\mathcal{F}^L$ and the global frame of prism measurements ($\mathcal{F}^1$, in the current case), for a sample $s_m$ in the \(j\)-th triplet.

The subsequent \(M\) poses form a distribution, of which we can extract an average translation and rotation defined by $\bm{\xi}_j \in \mathbb{R}^6$ in the global frame $\mathcal{F}^W$ for every $j \in \{1,J\}$.
The covariance $\bm{\Lambda}_j \in \mathbb{R}^{6\times6}$ of this distribution of poses yields the uncertainty on every vehicle pose $\bm{\xi}_j$.
The resulting reference trajectory is defined as the set $\xi$ of poses \(\left\{\bm{\xi}_1,\dots,\bm{\xi}_j,\dots,\bm{\xi}_J\right\}\).
$\Lambda$ is a set that contains the covariance $\bm{\Lambda}_j$ of every pose $\bm{\xi}_j$ along $\xi$.
These ground truth uncertainties can be used for the evaluation of six-\ac{DOF} trajectories.
In the next sections, we will characterize the impact of the source of noise over the uncertainty models of the prism trajectories and of the reference trajectory.

\section{Experiments}
\label{sec:experiments}

We used three Trimble S7 \acp{RTS} to track three Trimble MultiTrack Active Target MT1000 prisms with a measurement rate of \SI{2.5}{\hertz}.
The \autoref{tab:budget} gives the different kinds of noises that were modeled, in accordance with the specifications of the \textit{Trimble S7}.
Following the \ac{GUM} guidelines~\cite{GUM}, these noises have been divided into two types (\ie \textbf{A} and \textbf{B}).
The former is determined through experimental values (\eg extrinsic calibration, time synchronization error).
The latter is given by the specifications of the measuring instrument (\eg range, angle, tilt compensator), or from an environmental model (\ie atmospheric factors).

\vspace{+5mm}
\tabulinesep=0.6mm
\begin{table}[htbp]
    \centering
    \caption{\textit{Trimble S7} \ac{RTS} uncertainty parameters.
        These uncertainties have been categorized according to their type, source, distribution and values.
        Zero means are considered for each normal distribution, except for the time synchronization error as described in \autoref{sec:theoryA}.
        All standard deviation values are given for $2 \sigma$ according to the ISO norm ISO17123-3.
        The abbreviation $\text{ppm}$ means parts per million, and $\SI{1}{"}=\SI{4.85E-6}{\radian}$}
    ~\label{tab:budget}
    \begin{tabu}{X[1,c]X[3,l]X[1,c]X[3,c]}
        \toprule
        \multicolumn{2}{c}{\textbf{\emph{Influence factors}}} & \textbf{\emph{Distribution}} & \textbf{\emph{Values}} \\
        \midrule
        & \textbf{Extrinsic calibration}  &  & \\
        & - Translation                   & Normal & $\sigma_{tx}$, $\sigma_{ty}$, $\sigma_{tz}$  \\
        \textbf{Type A} & - Rotation                      & Normal & $\sigma_{rx}$, $\sigma_{ry}$, $\sigma_{rz}$  \\
        \cline{2-4}
        & \textbf{Time synchronization}   &  & \\
        & - Velocity                   & Normal & $\mu_{v}$, $\sigma_{v}$  \\
        & - Time error                 & Normal & $\mu_{t_s} = \SI{1.2}{\ms}$, $\sigma_{t_s}= \SI{0.8}{\ms}$ \\
        \midrule
        & \textbf{Instrument}  &  & \\
        & - Distances                     & Normal & $\sigma_{\rho}=\SI{4}{mm}+\SI{2}{ppm}$ \\
        & - Horizontal directions         & Normal & $\sigma_{\phi}=\SI{2}{"}$ \\
        & - Vertical directions           & Normal & $\sigma_{\theta}=\SI{2}{"}$ \\
        \cline{2-4}
        \textbf{Type B} & \textbf{Tilt compensator} &  & \\
        & - Angle bias              & Normal & $\sigma_{tilt}=\SI{0.5}{"}$ \\
        \cline{2-4}
        & \textbf{Atmospheric factors} &  & \\
        & - Temperature                & Uniform & $\sigma_{T}=[0,\SI{1}{\celsius}]$ \\
        & - Pressure                   & Uniform & $\sigma_{P}=[0,\SI{10}{\hecto\pascal}]$ \\
        & - Humidity                   & Uniform & $\sigma_{H}=[0,\SI{2}{\percent}]$ \\
        \bottomrule
    \end{tabu}
    \vspace{-2mm}
\end{table}

As shown in \autoref{fig:fig-intro}, all three prisms were mounted on a Clearpath Warthog \ac{UGV}.
A Robosense RS-32 and a XSens MTi-10 \ac{IMU} were used as part of an \ac{ICP}-based \ac{SLAM} framework, working at a rate of \SI{10}{\hertz}.\footnote{\url{https://github.com/norlab-ulaval/norlab_icp_mapper}}
The experiments were conducted from February 2022 to January 2023.
They include \num{20} deployments, of which \num{18} took place on the campus of Université Laval and two were done in the Montmorency research forest, \SI{75}{\kilo\metre} north of Quebec City.
These 20 deployments allowed us to conduct 48 experiments, for a total of \SI{50}{\kilo\metre} of \ac{RTS}-tracked prism trajectories.

The same procedure was applied during each experiment, in order to collect consistent and standardized data during the whole year.
Also, each deployment was completed by measuring accurately the position of the three prisms, rigidly installed on the robot, with a single \ac{RTS}.
These measurements are used as reference points to compute the inter-prism distances, as a way to control data for each experiment.
The point-to-point minimization method presented in \autoref{sec:theoryC} is also relying on these measurements.
Weather conditions and atmospheric values were obtained through the weather service of \textit{Environment and Climate Change Canada}.\footnote{\url{https://climate.weather.gc.ca/historical_data/search_historic_data_e.html}}

\section{Results}
\label{sec:results}

\subsection{Influence of the sources of uncertainty over the results}

We first evaluated the impact of different sources of noise on the prism position uncertainty.
These sources of noise are 1) the \ac{RTS} instrument noises, 2) the tilt compensator noises, 3) the atmospheric factors, 4) the time synchronization, and 5) the extrinsic calibration.
Every source of noise was represented by a distinct covariance matrix, on which we applied the Frobenius norm~\cite{Barfoot2014} to evaluate their effect on the prism position uncertainty.
These uncertainties were also compared for different ranges, to determine how it is impacted by the \ac{RTS}-prism distance.

\begin{figure}[htbp]
    \centering
    \includegraphics[width=\linewidth]{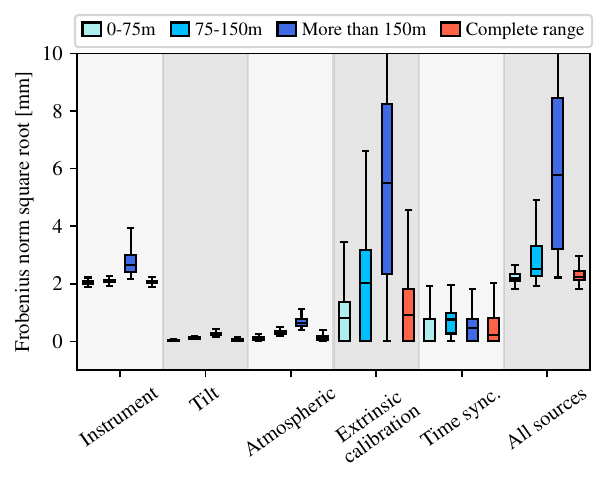}
    \caption{Influence of the noise sources from~\autoref{tab:budget}, in relation with the measured distance between an \ac{RTS} and its assigned prism.
    The square root of the Frobenius norm is used to estimate the similarity between covariance matrices.}
    \label{fig:influence_noise}
    \vspace{-2mm}
\end{figure}

\autoref{fig:influence_noise} shows that, for every noise source except the time synchronization noise, a longer range will lead to higher uncertainty.
This relation is especially the case for extrinsic calibration, with a median value of \SI{0.91}{\milli\metre} in the range of \num{0} to \SI{75}{\metre}, a median value of \SI{1.88}{\milli\metre} in the range of \num{75} to \SI{150}{\metre}, and a median value of \SI{5.49}{\milli\metre} in the range of more than \SI{150}{\metre}.
Moreover, this noise source affects the majority of the total uncertainty for the complete range, with a median value of \SI{1.32}{\milli\metre}.
This observation is coherent with the description of extrinsic calibration given in \autoref{sec:theoryA}, as this calibration relies on measurements that are all impacted by the other sources of uncertainty, causing its covariance to be higher.
However, this impact on the uncertainty is only considerable at a long-range, while the noises inherent to the \ac{RTS} have the highest median regardless of the distance.
With a median value of \SI{2.1}{\milli\metre}, we confirmed that the uncertainty level from the instrument is in the range of the manufacturer's specifications.
This noise also increases with long-range measurements, with a median value of \SI{2.78}{\milli\metre} for distances of more than \SI{150}{\metre}.
Meanwhile, the other sources of noise were less significant.
The time synchronization noise has a median value of \SI{0.4}{\milli\metre}, and does not depend on the measurement range.
Similarly, the atmospheric factors have a median value of \SI{0.16}{\milli\metre}, while the tilt noise has a median value of \SI{0.06}{\milli\metre}.
Both of these noise sources increase according to the measurement range.
In a field deployment, it would be important to keep in mind the two factors that have the highest influence on the results.
Therefore, it is crucial to achieve a good extrinsic calibration, as it is the main source of uncertainty for long-range measurements.
Otherwise, it is important to gather as much data as possible with ranges lower than \SI{150}{\metre}.
The median for all sources on the complete range is close to the median for shorter ranges, as we gathered more data at short distances than at long distances: \SI{80}{\%} of the data was taken with distances between \num{0} and \SI{75}{\metre}, \SI{14}{\%} between \num{75} and \SI{150}{\metre} and \SI{6}{\percent} for more than \SI{150}{\metre}.
Consequently, the results could be impaired by the lack of long-range measurements.
Overall, since the \ac{RTS} has an inherent noise, better results could be obtained with other instruments that would be more precise.


\subsection{Trajectories with uncertainty}

We used the pipeline from~\cite{Vaidis2022} to filter the raw prism measurements to increase the accuracy of the results.
The modules (\num{1} and \num{2}) from this pipeline were used with the parameters $\tau_r = \SI{2}{m.s^{-1}}$, $\tau_a = \tau_e = \SI{1}{deg.s^{-1}}$, $\tau_s = \SI{3}{s}$ and $\tau_l = \SI{2}{s}$.
Instead of using linear interpolation in the third module, we computed a \ac{GP} interpolation with the \ac{STEAM} library.
This \ac{GP} was used to interpolate the uncertainties from the \ac{MC} method, as explained in~\autoref{sec:theoryB}.

\begin{figure}[htbp]
    \centering
    \includegraphics[width=1\linewidth, trim={10 0 40 0}, clip]{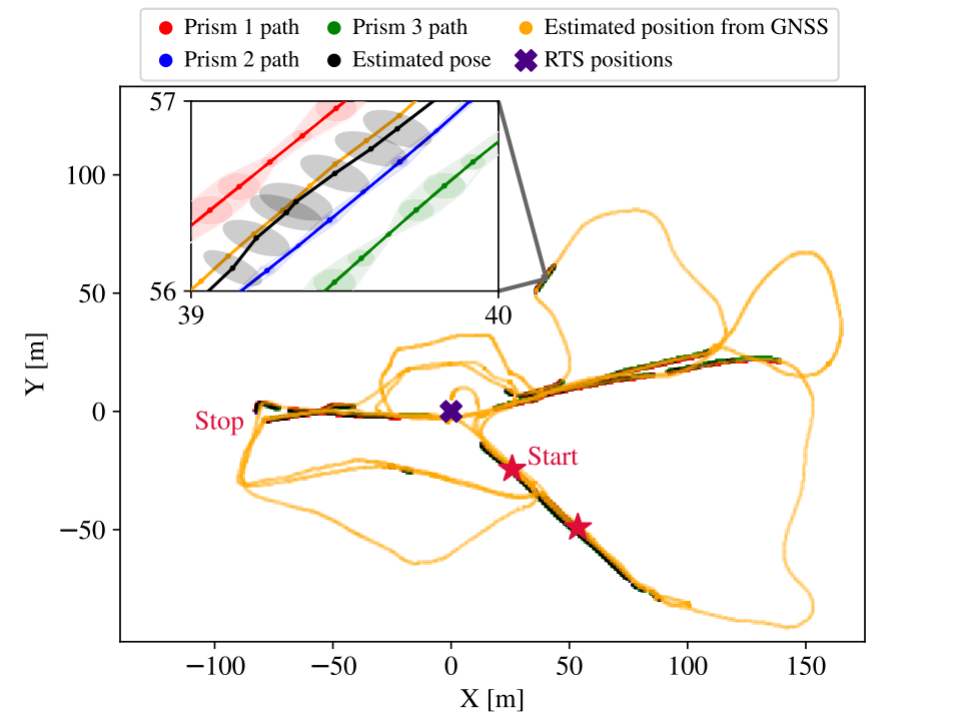}
    \caption{Top view of a reference trajectory generated from a deployment in the Montmorency forest.
    Interpolated prism paths $\widehat{\mathcal{Q}}^i$ are represented by dots in red (prism 1), blue (prism 2), and green (prism 3).
    The \ac{RTS}-estimated pose $\bm{\xi}$ is displayed by black points, along with a \ac{GNSS}-estimated position represented in orange.
    Both covariances $\widehat{\bm{\Sigma}}^i_j$ and $\bm{\Lambda}_j$ are shown with shaded ellipsoids.
    \acp{RTS} positions are indicated by an indigo cross, along with the start and stop positions indicated by red stars.}~\label{fig:trajectory}
    \vspace{-2mm}
\end{figure}

An example of this interpolation is shown in~\autoref{fig:trajectory}, which represents the results of a deployment at the Montmorency forest.
The interpolated prism measurements are displayed with red, blue, and green dots, along with their uncertainties as shaded ellipsoids.
The orange dots represent measurements from a \ac{GNSS} system on the robot that took data at a rate of \SI{5}{\Hz}.
As in the fourth module of the pipeline in~\cite{Vaidis2022}, the uncertainty has been filtered for values over \SI{20}{\centi\metre}, while the inter-prism distances are kept under \SI{10}{\centi\metre} to ensure that the values are precise enough for ground truth generation.
The point-to-point method described in~\autoref{sec:theoryC} propagates the prisms uncertainties among the reference pose of the Warthog, as shown with black dots in~\autoref{fig:trajectory}.
The six-\ac{DOF} pose and uncertainties on the ground truth trajectory can be compared with an estimated robot trajectory through the use of other metrics than the Euclidean norm.

Even if \ac{RTS} measurements are more accurate than \ac{GNSS} (\num{2}-\SI{3}{\cm}), they gather fewer data over time.
Therefore, with the \ac{GP} interpolation, the uncertainty on the \acp{RTS} measurements increases over time.
It can reach as much as \SI{5}{\cm}, as shown in the zoomed section of~\autoref{fig:trajectory}.
This issue can be solved by using \acp{RTS} with a higher measurement rate.
Moreover, the \ac{MC} method used with the point-to-point method spreads the error on the final robot pose.
Finally, as \acp{RTS} requires direct line-of-sight with a prism, fewer data can be measured in obstructed environments such as forests.
This constraint is visible on~\autoref{fig:trajectory}, where the \ac{RTS}-estimated poses only appear in areas with a direct line-of-sight from the \acp{RTS}.


\subsection{Impact of models over pose-uncertainty results}

\begin{figure}[htbp]
	\centering
	\includegraphics[width=\linewidth]{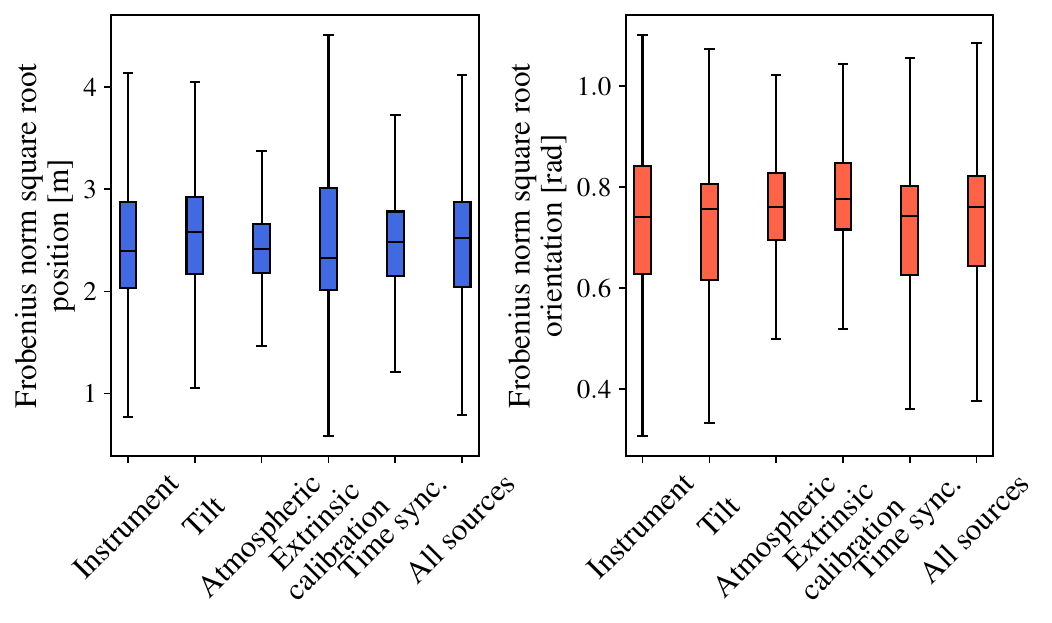}
	\caption{Impact of noise sources on the final pose uncertainty, for both the translation and the orientation, respectively in blue and red.}
	\label{fig:boxplot_pose}
    \vspace{-2mm}
\end{figure}

The~\autoref{fig:boxplot_pose} shows that the uncertainty on the position and orientation of a robot is not prominently affected by a single source of noise.
For instance, no matter the source of uncertainty, the medians are \SI{2.5}{\metre} and \SI{0.76}{\radian} for the position and the orientation, respectively.
This stability might come from the point-to-point minimization which smoothens the trajectory and therefore minimizes some of the errors that could be caused by the different sources of uncertainty (\ie uncertainty inherent to the \ac{RTS}, the tilt, the atmospheric conditions, the extrinsic calibration, and the time synchronization).

The values given by the Frobenius norm square root for the robot position (in ~\autoref{fig:boxplot_pose}) are larger by an order of magnitude than the uncertainty computed on a prism position (in ~\autoref{fig:influence_noise}).
This can be related to the \ac{GP} interpolation in the pipeline, as the interpolation drastically increases the uncertainty in proportion to the speed of the vehicle.
Also, the point-to-point minimization propagates the prism uncertainties on the vehicle pose uncertainty.
These results can be compared to the one obtained by \citet{Vaidis2021}, where they show the same kind of uncertainty on the final pose of a vehicle, with a comparable amount of uncertainty on the prism positions.

\section{Conclusion}
\label{sec:conclusion}
In this paper, we proposed a \ac{MC} method to model the uncertainties coming from multiple \acp{RTS} with the intent to better compare six-\ac{DOF} trajectories.
The estimated uncertainty of a prism measurement is then interpolated with a \acp{GP}, and propagated to the estimated six-\ac{DOF} pose of a robotic platform with a \ac{MC} method, used with a point-to-point minimization.
We have highlighted that the main source of noise when using multiple \acp{RTS} is coming from the extrinsic calibration, besides the uncertainty that is inherent to the instrument.
Our model has demonstrated that the uncertainty on a prism measurement is proportional to the distance between that prism and a \ac{RTS}.
Moreover, none of the sources of noise have a certain impact on the uncertainty of a pose that is computed with point-to-point minimization.
This can be caused by the minimization method that smoothens the effect of different noise sources to an average value.

Future works would include the optimization of our extrinsic calibration method to minimize resulting uncertainties.
Other atmospheric factors such as snow or rain would also need to be experimentally characterized.
The uncertainty of \acp{GNSS} could be modeled in the same manner to compare it with the uncertainty obtained with our method.
This would allow us to evaluate localization and mapping algorithms by merging information from \ac{RTS}-based ground truth trajectories with \ac{GNSS}-based ground-truth trajectories.


\section*{Acknowledgment}

This research was supported by the  Natural  Sciences and Engineering  Research  Council of  Canada  (NSERC)  through the grant CRDPJ 527642-18 SNOW (Self-driving Navigation Optimized for Winter).


\IEEEtriggeratref{6}
\IEEEtriggercmd{\enlargethispage{-0.1in}}

\printbibliography

\end{document}